\documentclass[10pt,twocolumn,letterpaper]{article}

\usepackage{cvpr}              % To produce the CAMERA-READY version
\definecolor{cvprblue}{rgb}{0.21,0.49,0.74}
\usepackage[pagebackref,breaklinks,colorlinks,allcolors=cvprblue]{hyperref}

\def\paperID{*****} % *** Enter the Paper ID here
\def\confName{CVPR}
\def\confYear{2026}

\title{MARS: Technical Report for the CASTLE Challenge at EgoVis 2026}

\author{
Haoyu Zhang$^{1\,2}$, Qiaohui Chu$^{1\,2}$, Yisen Feng$^{1}$, Meng Liu$^{3}$, Weili Guan$^{1}$, \\ Yaowei Wang$^{1\,2}$, Liqiang Nie$^{1}$\\
$^1$Harbin Institute of Technology (Shenzhen) \qquad  $^2$Pengcheng Laboratory    \\$^3$Shandong Jianzhu University\\
{\tt\small \{zhang.hy.2019, yisenfeng.hit, qiaohuichu8599, mengliu.sdu, honeyguan, nieliqiang\}@gmail.com;} \\ {\tt\small wangyw@pcl.ac.cn}
}

\begin{document}
\maketitle
\begin{abstract}
This report presents MARS, short for \textbf{M}ultimodal \textbf{A}gentic \textbf{R}easoning with \textbf{S}ource selection, our system for the CASTLE Challenge at EgoVis 2026. Participants must answer 185 closed-form questions over the CASTLE 2024 dataset. In contrast to prior single-video egocentric benchmarks, CASTLE requires reasoning over four days of activity, 15 synchronized perspectives, official transcripts, and multiple auxiliary modalities, including personal photos, auxiliary videos, gaze, thermal imagery, and heartrate measurements. MARS therefore treats the task as an agentic evidence-selection problem over multimodal sources rather than a purely text-only pipeline.
MARS first follows the official CASTLE directory organization to build evidence memories from two primary sources, videos and transcripts, and four auxiliary sources, gaze, heartrate, photos, and thermal imagery. Long videos are converted into captions and DeepSeek-based summaries only because CASTLE videos are too long to fit directly into the model context for every question; this step compresses temporal evidence while keeping photos and other auxiliary media available as source-specific evidence. At inference time, a GPT-5.4 decision agent repeatedly chooses whether to continue reasoning, request a specific missing modality, produce an answer, or fall back to a random option when the evidence remains insufficient. The resulting system achieved second place on the final CASTLE Challenge leaderboard. Our codes are available at \href{https://github.com/Hyu-Zhang/MARS}{https://github.com/Hyu-Zhang/MARS}.
\end{abstract}
    
\section{Introduction}

\begin{figure*}[t]
  \centering
   \includegraphics[width=\linewidth]{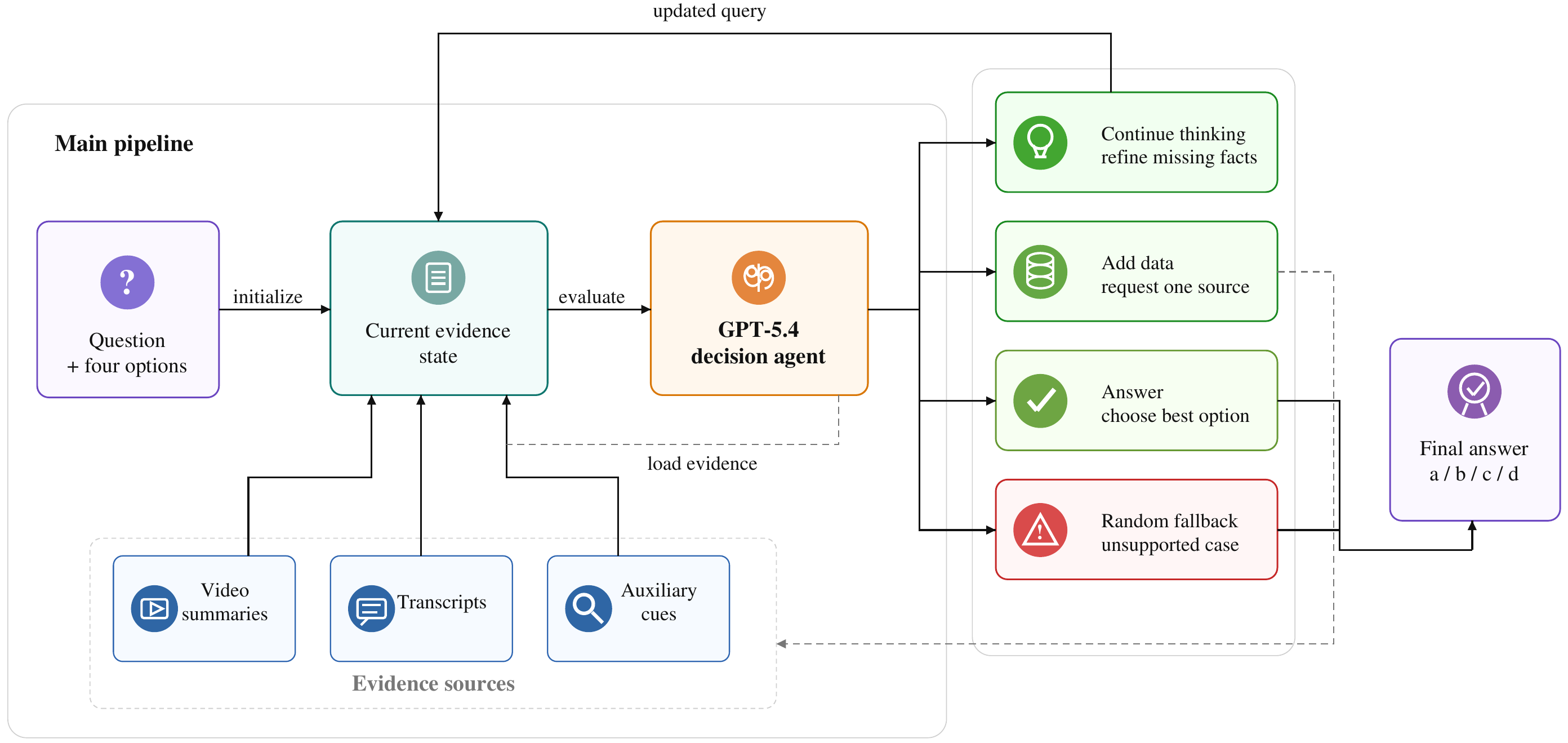}
\caption{Decision flow of MARS. The agent maintains a compact evidence state and repeatedly chooses one of four actions: continue thinking, request one additional source, answer, or use fallback when the sample remains unsupported.}
   \label{fig:method}
\end{figure*}

Egocentric understanding has moved from isolated clips to long-horizon, multi-session, and multi-sensor reasoning, including large-scale ego-exo data, long egocentric question answering, long-form video agents, exocentric-to-egocentric transfer, spatial video understanding, and long-term action anticipation~\cite{grauman2024egoexo4d,di2024grounded,wang2024videoagent,ren2024timechat,zhang2026exo2ego,song2024moviechat,zhang2026spatial,chu2026intention,chu2025technical}. The CASTLE 2024 dataset~\cite{rossetto2025castle} is a representative benchmark for this transition. It records four days of activities from 10 participants in a shared environment, yielding more than 600 hours of video from 15 perspectives together with speech, photos, gaze traces, thermal imagery, and biometric signals. Such a setting is substantially more realistic than short-form benchmarks, but it also makes question answering considerably harder: relevant evidence may appear in different rooms, at different times, and in different modalities.

Our solution also builds on prior studies on adaptive visual evidence selection, retrieval-augmented egocentric captioning, multimodal grounding, video-grounded dialog, identity cues, and multimedia retrieval~\cite{pmlr-v235-zhang24aj,xu2024retrieval,feng2024objectnlq,feng2025object,10.1145/3808220,zhang2021multimodal,10239469,guan2022bi,feng2025osgnet}. These works motivate a practical design principle for CASTLE: avoid processing all raw media at once, and instead select the evidence source that is most likely to resolve the current uncertainty.

The CASTLE Challenge @ EgoVis 2026 formulates this problem as closed-form question answering. Given the full dataset and a four-choice question, the system must select exactly one answer. The official test set contains 185 questions, and evaluation is based solely on accuracy. Many questions require more than object recognition. They demand person identification, day-level temporal localization, cross-view verification, OCR, and transcript-grounded reasoning. The official release is organized around a \texttt{main/} directory with \texttt{day1}--\texttt{day4} subdirectories, and an \texttt{auxiliary/} directory containing \texttt{gaze}, \texttt{heartrate}, \texttt{photo}, \texttt{thermal}, and \texttt{video} data.\footnote{\url{https://huggingface.co/datasets/CASTLE-Dataset/CASTLE2024/tree/main}} This structure naturally suggests a source-aware reasoning system rather than a single monolithic video model.

Instead of training a specialized multimodal model from scratch, we adopt a pragmatic source-selection strategy derived from HCQA~\cite{zhang2024hcqa}. We name this adaptation MARS, standing for \emph{Multimodal Agentic Reasoning with Source selection}. The key idea is to maintain a compact evidence state and let a capable decision model iteratively choose which source should be used next. Video is summarized into text because the raw CASTLE videos are long and cannot be repeatedly inserted into the context window; by contrast, photos, gaze, heartrate, and thermal evidence are treated as auxiliary modalities that can be requested when they are useful for the current question.

Based on this observation, MARS uses an agentic loop for CASTLE. The system starts from the question and answer options, consults the primary evidence built from video summaries and transcripts, and then asks the decision model whether the current evidence is sufficient. If not, the agent chooses which additional modality should be loaded next, such as gaze for attention cues, photos for identity or object grounding, thermal images for heat-related events, or heartrate for activity intensity. The loop ends when the model has enough evidence to answer, reaches the reasoning budget, or explicitly judges that the sample is unsupported. This simple adaptation transfers well from last year's framework to the new benchmark and ultimately ranks second on the final leaderboard.

\section{Methodology}

\begin{table*}[ht]
  \caption{Solution evolution of MARS during the CASTLE Challenge.}
  \centering
  \small
  \begin{tabular}{ccp{0.66\textwidth}}
    \toprule
    \textbf{Order} & \textbf{Accuracy} & \textbf{Modification} \\
    \midrule
    1 & 0.35 & Zero-shot question answering using only the question text and answer options. \\
    2 & 0.42 (+0.07$\uparrow$) & Inject official transcripts from likely streams, enabling transcript-grounded answering. \\
    3 & 0.44 (+0.02$\uparrow$) & Add person, day, room, and event cue parsing to narrow retrieval to relevant CASTLE directories. \\
    4 & 0.48 (+0.04$\uparrow$) & Convert retrieved video clips into HCQA-style captions, OCR notes, and DeepSeek summaries. \\
    5 & 0.51 (+0.03$\uparrow$) & Add source-specific auxiliary evidence from photos, gaze, heartrate, thermal images, and auxiliary videos. \\
    6 & 0.55 (+0.04$\uparrow$) & Introduce a GPT-5.4 decision agent that can continue thinking, request additional data, or answer. \\
    7 & 0.57 (+0.02$\uparrow$) & Add final consistency checking and unsupported-case random fallback under a fixed reasoning budget. \\
    \bottomrule
  \end{tabular}
  \label{tab:evolution}
\end{table*}

\begin{table}[ht]
  \caption{Final leaderboard of the CASTLE Challenge.}
  \centering
  \begin{tabular}{ccc}
    \toprule
    \textbf{Rank} & \textbf{Participant} & \textbf{Accuracy} \\
    \midrule
    1 & WDL & 0.58 \\
    \textbf{2} & \textbf{ilearn\_zhy (ours)} & \textbf{0.57} \\
    3 & raghad\_khaled & 0.55 \\
    4 & SB-CuriosAI & 0.50 \\
    5 & SB-CuriosAI & 0.35 \\
    6 & vitality & 0.21 \\
    \bottomrule
  \end{tabular}
  \label{tab:result}
\end{table}

Our method, MARS, keeps the original HCQA spirit for long-video compression while extending it into a multimodal source-selection agent, as shown in Figure~\ref{fig:method}. The system does not assume that all evidence must be text. Instead, it maintains a compact evidence state and lets the agent decide which source to read next from a four-day, multi-view, multi-modal directory.

\subsection{Data Organization and Source Taxonomy}

We follow the official CASTLE release layout. The \texttt{main/} directory is organized by \texttt{day1}--\texttt{day4}; each day contains participant egocentric streams and static room views. The \texttt{auxiliary/} directory contains additional \texttt{gaze}, \texttt{heartrate}, \texttt{photo}, \texttt{thermal}, and \texttt{video} folders. In our system, these files are grouped into two primary sources and four auxiliary sources.

\textbf{Primary sources.} The first primary source is video. It includes the main egocentric and static streams as well as auxiliary videos when they are relevant to the question. We process videos offline following the HCQA pipeline~\cite{zhang2024hcqa}: long videos are sliced into short clips, representative frames are captioned, OCR is applied when screens or text appear, and clip-level captions are compressed into event summaries with DeepSeek~\cite{guo2025deepseek}. This conversion is used because CASTLE contains hundreds of hours of video, making direct raw-video prompting infeasible under context and cost limits. The second primary source is transcript. Official transcripts are kept as plain text, indexed by day, stream, and time segment, and can be injected into the reasoning context without visual preprocessing.

\textbf{Auxiliary sources.} Gaze files provide attention cues, especially when a question asks what a participant looked at or interacted with. Heartrate files provide weak but useful activity-intensity evidence. Photos provide high-resolution identity, object, vehicle, food, and screen cues. Thermal images provide heat-related evidence for cooking, appliances, drinks, or environmental state. These sources are not always necessary, but they are valuable when the primary video-transcript evidence is ambiguous.

\subsection{Offline Evidence Memory Construction}

Before answering questions, we convert reusable sources into an evidence memory. For each video stream, we sample clips at a fixed temporal stride and generate captions that emphasize people, objects, actions, text on screens, and changes in state. Adjacent captions are then summarized into compact time-window notes. This follows the HCQA idea of replacing expensive long-video reasoning with a searchable language representation, but adapts it to CASTLE by indexing notes with day, participant or room, timestamp, and modality.

Transcripts are treated differently because they are already textual. We normalize them into short utterance windows and preserve time information so that transcript snippets can be aligned with nearby video summaries. Auxiliary media are kept as source-specific evidence and summarized only when useful for reasoning: photo notes describe static visual details, thermal notes describe visible heat patterns, gaze notes describe likely attended targets when available, and heartrate notes describe local increases, decreases, or abnormal peaks. The result is a unified evidence table in which every entry has the same high-level schema: source type, day, stream or owner, time if available, and evidence content.

\subsection{Agentic Decision Loop}

At inference time, MARS receives the question, four answer options, and a small initial evidence package retrieved from video summaries and transcripts. We use GPT-5.4 as the decision model. Rather than forcing a one-shot prediction, the model is prompted to choose one action from a fixed action space:

\begin{itemize}
    \item \textbf{Continue thinking}: reorganize the current evidence, identify missing facts, and refine the retrieval query without loading new data.
    \item \textbf{Add data}: request one specific source, such as transcript, video summary, gaze, heartrate, photo, or thermal evidence, together with the target day, person, room, or time range.
    \item \textbf{Answer}: select one option when the current evidence is sufficient and cite the supporting evidence in the internal rationale.
    \item \textbf{Random fallback}: if the reasoning budget is exhausted and the evidence remains insufficient, randomly choose one option because the benchmark requires a closed-form answer.
\end{itemize}

This loop is important because different CASTLE questions require different evidence. A question about a spoken explanation can often be solved from transcripts alone. A question about a car charging screen or a cooking state requires visual captions and OCR. A question about what a person attended to may benefit from gaze, while a question involving physical effort may benefit from heartrate. The agent therefore acts as a source controller: it reads cheap textual evidence first and only adds auxiliary modalities when the current state exposes a concrete need.

\subsection{Answer Generation}

When the agent selects \textbf{Answer}, we run a final consistency check over the accumulated evidence. The decision model compares every option against the evidence table and marks unsupported options before returning the final label. For low-confidence cases, we repeat the loop with a stricter prompt that asks the agent to find the single missing evidence type most likely to distinguish the top two options. If no such evidence is available, the system uses the random fallback action and records the sample as unsupported.

This design makes the pipeline more robust than a fixed retrieval recipe. It preserves the simplicity of HCQA for long-video compression, uses DeepSeek summaries to keep video evidence compact, and lets GPT-5.4 dynamically decide when transcripts are enough and when auxiliary modalities should be added.

\section{Experiment}

\subsection{Performance Comparison}

The official competition released 185 test questions and evaluated submissions using accuracy. As shown in Table~\ref{tab:result}, our submission under the participant name \emph{ilearn\_zhy} achieved 0.57 accuracy and ranked second on the final leaderboard. This result is noteworthy because CASTLE is substantially harder than standard single-video QA benchmarks: each question may require evidence retrieval across days, identities, rooms, and modalities rather than within a short self-contained clip.

\section{Solution Evolution}

Table~\ref{tab:evolution} summarizes how the final system was formed. The initial solution started from a weak text-only baseline. The first large improvement came from using the official transcripts, which immediately introduced event and speech cues absent from the raw question. The next gains came from respecting the official directory structure: once we identified the relevant day, participant, room, or auxiliary owner, we could retrieve a much smaller evidence set and summarize it into short notes. HCQA-style video captions and DeepSeek summaries then made long video evidence searchable under context limits, while auxiliary photos, gaze, heartrate, thermal images, and auxiliary videos improved grounding for cases that could not be solved from transcripts alone.

The final transition from 0.55 to 0.57 came from the full MARS design. Instead of trusting a fixed retrieval recipe, we let GPT-5.4 decide whether to continue reasoning, request a missing modality, answer, or use the required random fallback for unsupported cases. This design was particularly useful for questions involving numbers, temperatures, colors, counts, attention, or physical state, where one weak transcript snippet or one missed visual detail could otherwise change the final choice.

\section{Conclusion}

We presented MARS, our submission to the CASTLE Challenge at EgoVis 2026. By preserving the original HCQA idea of compressing long videos into searchable summaries while allowing the agent to request non-video auxiliary sources, MARS adapts naturally to CASTLE's four-day, multi-stream, and multi-modal setting. The final method achieved 0.57 accuracy and ranked second on the official leaderboard.

These results indicate that a source-selective multimodal agent remains a strong and practical solution for long-horizon egocentric question answering. At the same time, the observed gains across the solution evolution process show that performance depends heavily on better evidence indexing, stronger cross-view grounding, and more reliable source-selection policies.

{
    \small
    \bibliographystyle{ieeenat_fullname}
    \bibliography{main}
}


\end{document}